\documentclass[letterpaper, 10 pt, conference]{ieeeconf}  
\usepackage{graphicx}
\usepackage{algorithm}
\usepackage{amsmath}
\usepackage{amssymb}
\usepackage{algpseudocode}
\usepackage{tikz}
\usepackage{pgf-pie}  
\usetikzlibrary{shapes,positioning}
\usepackage{etoolbox}
\usepackage{afterpage,lipsum}
\usepackage{cite}
\usepackage{bm}
\usepackage{subcaption} 

\title{\LARGE \bf Soil Sample Search in Partially Observable Environments}
\IEEEoverridecommandlockouts
\author{Han~Yang$^1$, Andrew~Dudash$^2$
\thanks{$^1$ Noblis Autonomous Systems Research Center Reston, Virginia \tt{\small Han.Yang@noblis.org}}%
\thanks{$^2$ Noblis Autonomous Systems Research Center Reston, Virginia
\tt{\small Andrew.Dudash@noblis.org}}%
}

\begin{document}

\maketitle
\thispagestyle{empty}
\pagestyle{empty}

\begin{abstract}
To work in unknown outdoor environments, autonomous sampling machines need the ability to target samples despite limited visibility and robotic arm reach distance. We design a heuristic guided search method to speed up the search process and more efficiently localize the approximate center of soil regions. Through simulation experiments, we assess the effectiveness of the proposed algorithm and discover superior performance in terms of speed, distance traveled, and success rate compared to naive baselines.
\end{abstract}
\vspace{+10pt}

\section{INTRODUCTION}
\label{sec:introduction}
In this paper, we address the problem of autonomous sample collection in outdoor, unknown environments. While collecting soil or similar organic material, there are no guarantees that samples will be reachable, visible, or even present. For this reason, a robot needs an effective search task to locate and move sufficiently close to the samples prior to beginning collection. We formulate this problem as a target search problem for one of multiple irregularly shaped, i.e. amorphous, target regions in a heterogeneous terrain, similar to how stretches of open soil appear in nature.

Specifically, we employ a heuristic guided method to locate the centers of soil clusters containing minimum non-pickable obstacles, like rocks or grass. This search method assumes the camera provides a bird's eye field of view, but can be applied to a variety of sample collection applications: an unmanned aerial vehicle that identifies and reports large regions containing samples or the vision system of a unmanned ground vehicle moving to pick an obstacle. While this method may not be optimal, it presents a simple to implement algorithm that requires no prior knowledge about target distribution, and has experimentally been found effective and robust against visibility impairments due to factors such as fog and dust. The contributions of this paper are as follows:
\begin{itemize}
    \item We develop a heuristic that allows autonomous robots to find the center of an amorphous target in a 2D environment, with applications to soil and organic material collection in heterogeneous outdoor terrains.   
    \item We experimentally measure the effectiveness of our heuristic method compared to a variation on it and two offline-generated search patterns. Our experiments included over 5000 trials across different soil abundance levels and visibility conditions.
\end{itemize}
\afterpage{
    \begin{figure}[t!]
        \centering
        \includegraphics[width=0.3\textwidth]{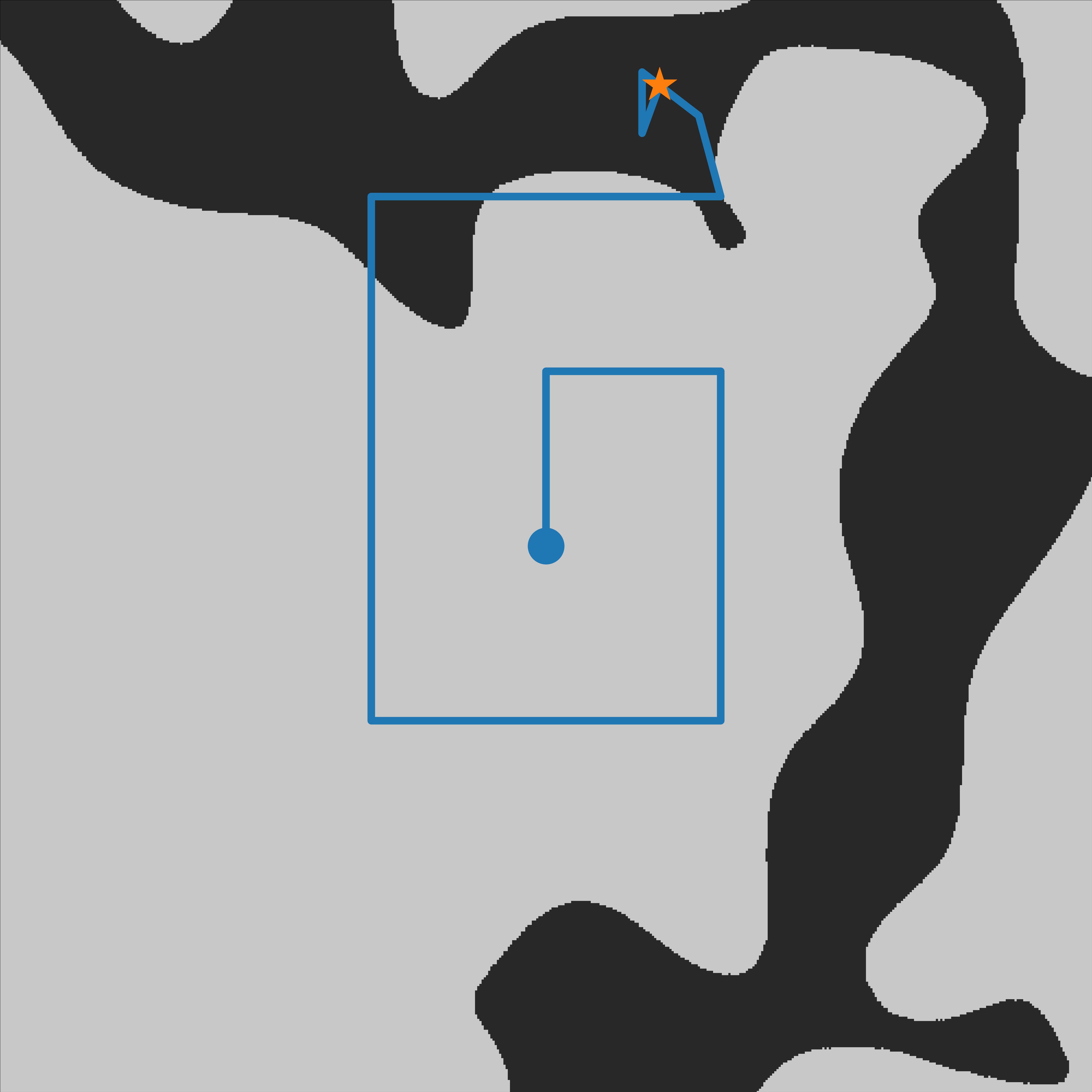}
        \caption{ In this example, a robot---perhaps a camera mounted end effector of a robotic arm---uses a heuristic method to search for the center of a soil region in a sample distribution. The circle is the start position, and the star indicates the center which the agent aims to reach.}
        \label{fig:intro_figure}
        \vspace{-18pt}
    \end{figure}
}

\section{RELATED WORK}
\label{subsec:related work}
The rapid development of autonomous robotics has drastically improved the efficiency and success rate for target search and tracking problems due to improved maneuverability, versatility, and ability to make decisions with limited information. Such systems have been applied to a multitude of fields including home services \cite{Hsu_Wee_Sun_Lee_Rong_2008}, marine maintenance \cite{Meghjani_Manjanna_Dudek_2016}, field exploration \cite{Zhao_Yan_Xie_Dai_Wei_2024}, and wilderness search and rescue (SAR)\cite{Wu_Ju_Wu_Lin_Xiong_Xu_Li_Liang_2019, Goodrich_2008}. 

Searching for a single, static object is the simplest form of the target search problem. For this case, a popular class of solutions discretizes the search domain through methods such as common grid-based cellular decomposition \cite{Gupta_Bessonov_Li_2017} and graph-based representation. \cite{Botao_Jun_Qiang_Chepinskiy_2018}. Compared to continuous state space methods, these approaches simplify optimal solution computation. For example, with discretization, it is easier to formulate solutions to dynamic target problems---like partially observable Markov decision processes (POMDP) \cite{Hsu_Wee_Sun_Lee_Rong_2008}---because the motion model is also simplified. Moreover, a paper in frontier exploration \cite{Kim_Gupta_Sentis_2021, Yamauchi_1997} decompose the search region into square grids to simplify frontier and entropy map representation. Nevertheless, in our problem with amorphous target regions, the discretization approach may result in partially filled grids that are difficult to classify, as well as multiple fully filled grids that cannot be tackled with single target search approaches.  

Another popular class of solutions represents the problem using a continuous state space as studied by \cite{Wei_Huang_Lu_Song_2019}, \cite{Bhagat_Sujit_2020}. Both papers use deep Q-network (DQN) designs compatible with continuous state spaces for UAV SAR and surveillance tasks. In problems with non-trivial low-level dynamics, adopting a continuous solution helps to ensure the dynamic feasibility of planned paths \cite{Li_Wu_2020}. Still, these approaches focus on a single point-based target whose state can easily be represented by a small number of coordinates, which can be challenging to extend into problems involving amorphous targets. 

Many papers have generalized the search problem to locating multiple targets. When the number of targets is known in advance, RL and MDP based approaches can efficiently locate all targets that are either static or dynamic \cite{Wu_Ju_Wu_Lin_Xiong_Xu_Li_Liang_2019, Meghjani_Manjanna_Dudek_2016} with little online computation time. Luo et al. and Vanegas et al. studied the multi-target search problem with static obstacles in the search region to better extend the solution to outdoor search missions \cite{Luo_Zhuang_Pan_Feng_Shen_Gao_Cheng_Zhou_2024, Vanegas_Campbell_Eich_Gonzalez_2016}. While such approaches are extremely time efficient, they require priori knowledge on the target's distribution and do not scale well to problems without a well-defined number of targets. 

Coverage path planning (CPP) is a natural solution to locate all the targets in the search region without information regarding target quantities. A prominent class of such approach involves offline, deterministic path planning. An example is the lawnmower search pattern, which has been officially standardized for marine-time search and rescue \cite{IAMSAR_2022}. While such methods can guarantee success in most cases, they are often less efficient and cannot be naively employed without priori information about the search space. In this regard, Ianenko et. al.  \cite{Ianenko_Artamonov_Sarapulov_Safaraleev_Bogomolov_Noh_2020} and Champagnie et. al. \cite{Champagnie_Arvin_Hu_2024} developed proximal policy optimization and multi-agent entropy maximization based CPP approaches that more efficiently cover an unknown environment with dynamic and static obstacles. 
 
In our problem, the objective is to locate a singular target region, and full coverage approaches that search for every possible target in the domain may lead to unnecessary effort. In addition, most approaches assume point-based target models, which may fail to capture the amorphous nature of our search targets. Although treating the target areas as clusters of points may work in conjunction with search techniques developed for an unknown number of targets, for example, that detailed in \cite{Yousuf_2022}, it does not exploit clustering information to improve search efficiency. Therefore, we seek a solution that accounts for the amorphous and clustered nature of soil beds. 

Lastly, unlike many search path planning methods that assumes a fixed field of view \cite{Wu_Ju_Wu_Lin_Xiong_Xu_Li_Liang_2019, Meghjani_Manjanna_Dudek_2016, Ianenko_Artamonov_Sarapulov_Safaraleev_Bogomolov_Noh_2020, Champagnie_Arvin_Hu_2024}, we leverage extra information that can be gathered by simply increasing the camera's height and thus the view size, similar to methods studied by Gupta et. al., Bhagat et. al., Goodrich et. al \cite{Gupta_Bessonov_Li_2017, Bhagat_Sujit_2020, Goodrich_2008}. In particular, Gupta and Goodrich each assume that the resolution of the camera's view scales linearly with height. However, in our implementation, we consider other factors such as fog and dust which are prevalent in outdoor environments and may further impair visibility. 

\section{APPROACH}
\label{sec:approach}

\subsection*{Overview}
\label{subsec:overview}
In our problem formulation, we make simplifying assumptions that state transitions and position feedback are perfect and that the search region is open and free of obstacles. The state space $\mathbb S$ of the robot is a 3 tuple containing the possible positions of the agent $p = (x_A, y_A, z_A)$. Given limitations on the camera's resolution and physical bounds of a robot's workspace, we constrain the robot's height to be between $z_A \in \left[h_{\min{}},h_{\max{}}\right]$ and the location $(x_A, y_A)$ to be within some bounds $x_A \in \left[x_{\min{}},x_{\max{}}\right], y_A \in \left[y_{\min{}},y_{\max{}}\right]$. In addition, the observation space $\mathbb O$ includes the target's location and the robot's view. At each time step, the robot receives a perfect measurement of its position $p$ and fixed size, noisy image $\Tilde{I}$. Similar to sampling methods implemented by \cite{Dudash_Andrades_Rubel_Goli_Clark_Ewald}, we make the image a grayscale segmentation plot that represents whether a pixel contains soil samples or not. For simplicity, we convert the segmented image into a binary valued matrix, $I_b$, given some threshold $t$, represented by Equation \ref{eq:threshold}.
\begin{equation}
    \label{eq:threshold}
    I_{b,(i,j)} = \left\{\begin{aligned}
        &0,\quad \Tilde{I}_{(i,j)} < t,\\
        &1,\quad \Tilde{I}_{(i,j)} \geq t.
    \end{aligned}\right.
\end{equation}

To leverage the variable zoom size that can be achieved by our system, we include an action in the $z$ direction such that the agent can scan a larger or smaller area when needed. Therefore, we design the action space $\mathbb A$ to be a 3 tuple $a = (\Delta x, \Delta y, \Delta z)$, corresponding to changes in the agent's position.

\subsection*{Heuristic Policy}
\begin{figure}
    \centering
    \includegraphics[width=\columnwidth]{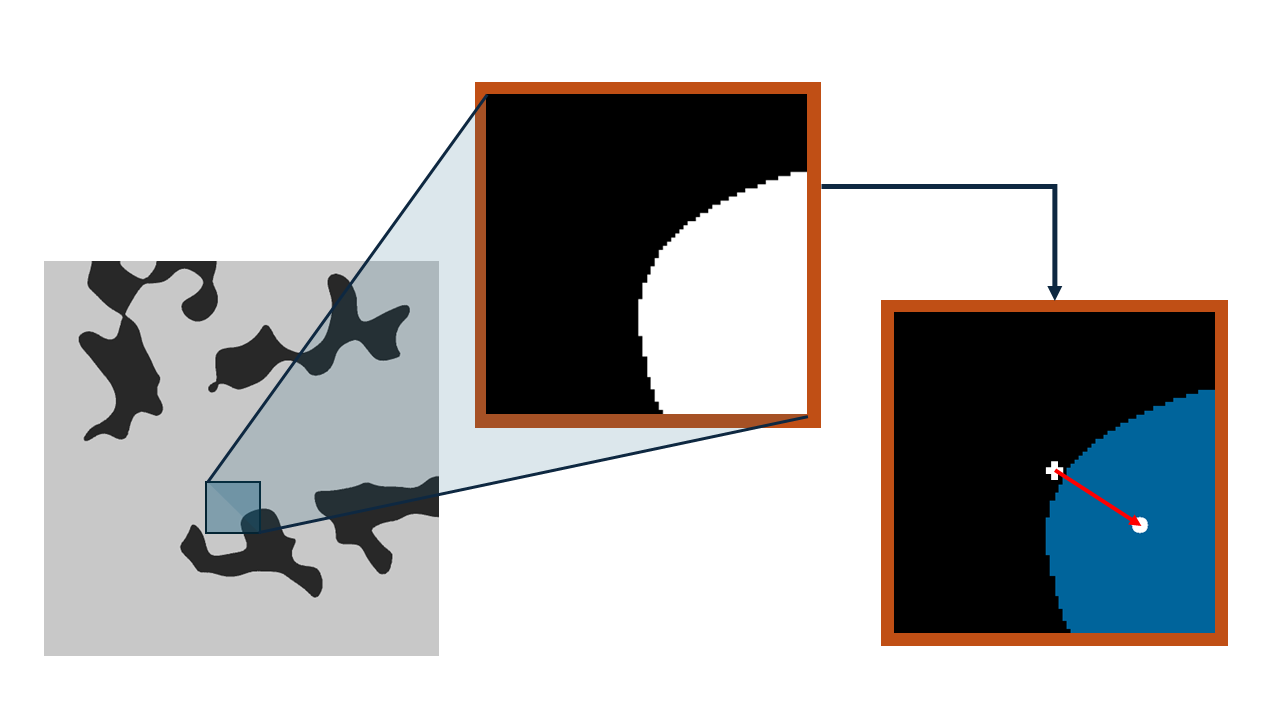}
    \caption{In this illustration, the contour of the spotted soil samples are shown as the blue region in the rightmost image. The centroid, indicated by the white dot, is determined from the contour and used to define the next waypoint and action, shown by the red arrow.}
    \vspace{-15pt}
    \label{fig:heuristic}
\end{figure}

Our search strategy follows a heuristic on the soil cluster's center given the current observation, as shown in Figure \ref{fig:heuristic}. At each time step, given a binary valued image $I_b$, we first compute the point and direction towards which the agent would move. This point is defined as the center of soil samples viewed by the robot and is computed by tracing the contours of observed soil clusters, then computing the centroid of the largest contour using Equation \ref{eq:centroid},
\begin{align}
    \label{eq:centroid}
    A\Bar{x} &= \overset{n}{\underset{x=0}{\sum}}\,\overset{n}{\underset{y=0}{\sum}}\, x I_{c,(y,x)}\\
    A\Bar{y} &= \overset{n}{\underset{y=0}{\sum}}\,\overset{n}{\underset{x=0}{\sum}}\, y I_{c,(y,x)},
\end{align}
where $A$ is the area of the largest contour, $I_c$ is a copy of $I_b$ with only the largest contour, and $(\bar x, \bar y)$ is the centroid in image coordinates. Afterwards, we take a greedy step by assuming that the center of the soil region is in the direction of the computed centroid. Therefore, the locally optimal solution would be to move in the relative direction between the current location, which is the center of the image frame, and the centroid, given by Equation \ref{eq:action}.
\begin{align}
    \label{eq:action}
    \Delta x &= \gamma(\Bar x - x_A)\\
    \Delta y &= \gamma(\Bar y - y_A),
\end{align}
where $\gamma$ is a scalar zoom size dependant on $z_A$ to convert the step into world coordinates. In addition to traversing in the 2D plane, once soil samples are spotted, the robot would lower its height to increase the observation accuracy, similar to the logic of a typical human pilot in SAR operations. 

\label{subsec:search}

In situations where no soil samples are spotted, the agent searches for targets through exploration. The notion of exploration here applies not only to exploring un-visited regions in the 2D plane, but also descending in height to further verify whether the current location contains soil samples. In the first case, labeled as $A_\text{explore}$, we program the agent to follow a deterministic baseline path, such as the expanding square pattern described in \cite{IAMSAR_2022}, while ascending for a larger viewing area. In the latter case $A_\text{dive}$, the robot would descend vertically without moving along the x-y plane. 

To balance between trusting the observation or not, the agent chooses either one of the two actions in a probabilistic manner. The decision probability depends on the expected accuracy of the received observation, given by Equation \ref{eq:transition}.
\begin{align}
    \label{eq:transition}
    &P(A_\text{explore}) = f(\hat{z}_A, \theta)\\
    &P(A_\text{dive}) = 1-f(\hat{z}_A, \theta),
\end{align}
where $\hat{z}_A$ is the normalized height, and $\theta$ represent visibility parameters. While the true relation between observation accuracy, visibility, and height is unknown, for simplicity, we define the model to be parameterized by a scalar visibility factor and set the decision probability given by Equation \ref{eq:decision_prob}. 
\begin{align}
    \label{eq:decision_prob}
    f(\hat{z}_A, \theta) = \frac{e^{\hat{z}_A/\theta}}{e^{\hat{z}_A/\theta} + 1}.
\end{align}

\begin{algorithm}[h]
  \caption{Heuristic Search Method}\label{alg:cap}
  \begin{algorithmic}
    \While{not in soil region}
        \State receive observation $o = \{p, I_b\}$ 
        \If{no target spotted}
            \State compute $\hat{z}_A$ and $f(\hat{z}_A, \theta)$
            \State descend with probability $f(\hat{z}_A, \theta)$
            \State follow pattern with probability $1-f(\hat{z}_A, \theta)$
        \Else
            \State find contours from $I_b$ 
            \State find centroid $(\bar x, \bar y)$ of the largest contour 
            \State move towards centroid and ascend
        \EndIf
        \State action = $(\Delta x, \Delta y, \Delta z)$
    \EndWhile
  \end{algorithmic}
\end{algorithm}

\section{EXPERIMENTS}
\label{sec:experiments}
We validate our search strategy in a simulated 2D environment with a randomized distribution of soil samples and a static initial position for our agent, similar to the setup in \cite{Meghjani_Manjanna_Dudek_2016}. We measure and compare the step count, distance traveled, and success rate of four different search methods under two different visibility scenarios. 

\subsection*{Simulation}
We developed a simulation environment using the Gymnasium package in Python \cite{towers_gymnasium_2023}. The environment is saved as a 1000x1000 image, with each pixel occupying a binary value that indicates whether the pixel contains a soil sample.

\begin{figure}[h]
\centering
\includegraphics[width=0.5\columnwidth]{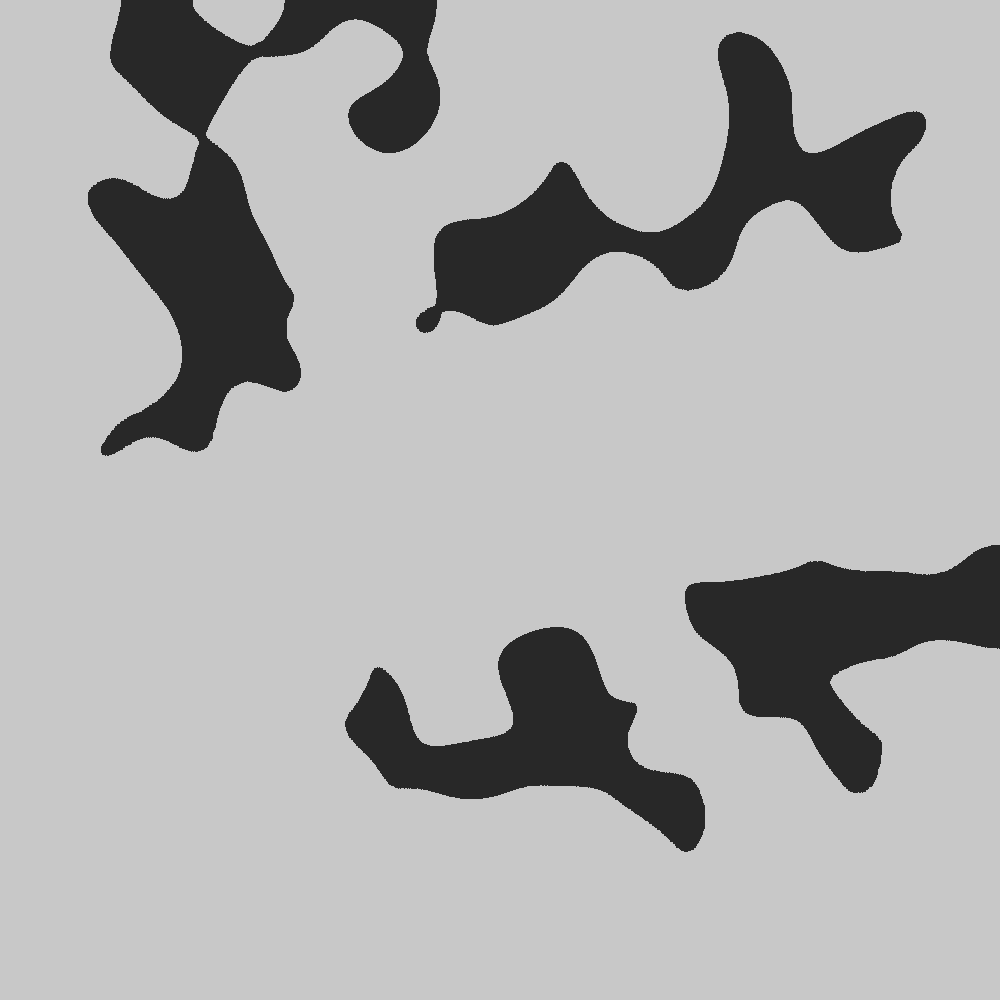}
\caption{In this example, a simulation environment with an amorphous soil distribution is shown. The dark regions represent soil, that can be sampled, and the lighter regions represent all other materials.}
\label{fig:sim_env}
\end{figure}

We utilize the OpenCV library \cite{opencv_library} to generate amorphous and random target areas. Specifically, we start by generating a white noise image that is then transformed into rough contours with the \verb|GaussianBlur| method. The outline of the contours is then highlighted by re-scaling the intensity and applying a binary threshold. Lastly, holes in the target areas are closed through the \verb|morphologyEx| method. To avoid the trivial case, we do not spawn the robot within a target region. An illustration of the generated environment is shown in Figure \ref{fig:sim_env}.

For each step of the experiment, the search agent outputs an action to the environment, after which the environment outputs observation information. The observation contains the current location, in pixels, of the robot, the robot's view as a 25x25 image, the distance traveled, and the step count, which will later be used as metrics for search performance. 

\begin{figure}[h]
\centering
\includegraphics[width=\columnwidth]{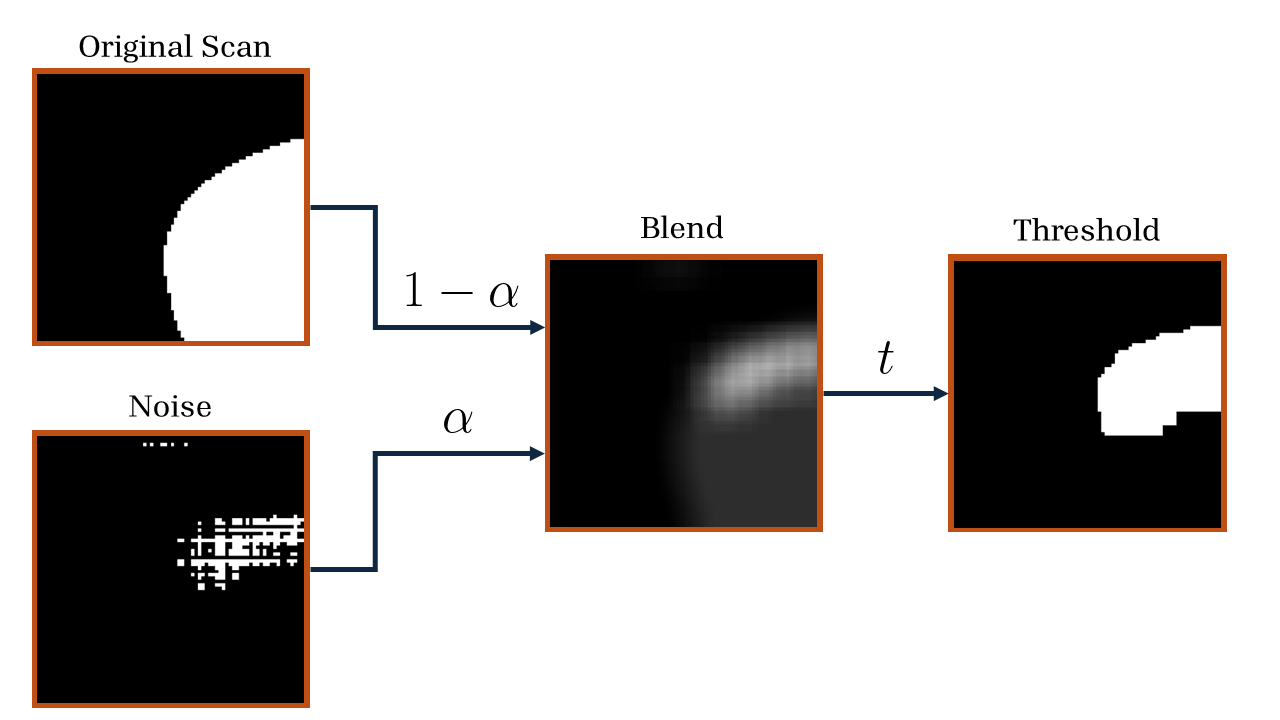}
\caption{To artificially impair the robot's visibility, we add noise, weighted by $\alpha$, to the ground truth view. The blended image is then passed through a threshold to obtain a binary valued image. In this example, parts of the original soil region are hidden from the agent by noise. }
\label{fig:noise} 
\end{figure}

Note that although the size of the robot's view size changes with respect to height, much like a real camera, the search policy expects images with fixed dimensions, thus all observations are resized before being passed to the agent. In addition, to simulate sensor noises and visibility impairment, we add artificial noise to the robot's view using OpenCV's \verb|addWeighted| method. Given that impaired visibility would generally result in false negative detection (failure to detect soil samples), the noise added is set to be predominantly 0, as shown in Figure \ref{fig:noise}. 

Similar to Equation \ref{eq:decision_prob}, the robot's visibility is parameterized by a single parameter that dictates the shape of a sigmoid blending function, given by Equation \ref{eq:height}. 
\begin{align}
    \label{eq:height}
    \alpha = \frac{e^{\hat{z}_A/\theta}}{e^{\hat{z}_A/\theta} + 1} + \nu,
\end{align}
where $\alpha$ is the weight of the noisy image, $\nu$ is a normal random scalar with mean of 0 and variance of 0.15.

Each trial starts with the robot spawning at the center of the map with a height of 50\% of $h_{\max{}}$, and terminates under one of three conditions:
\begin{itemize}
    \item \textit{Target Found}: The robot identifies the central region of one of the clusters with high certainty, quantified by a scan with above 95\% soil and the robot at height below 10\%.
    \item \textit{Truncated}: 
    The step number exceeds 300.
    \item \textit{Out of Bounds}: The robot attempts to move to a point outside of the 1000x1000 environment.
\end{itemize}

We implemented four search methods, two baselines and two versions of our heuristic method. We use the baselines to benchmark the performance of our proposed algorithm. Each method is detailed below: 
\begin{itemize}
    \item \textit{Expanding Square}: 
    The expanding square is a deterministic search pattern suggested by \cite{IAMSAR_2022} and is commonly used in search and rescue operations. 
    
    \item \textit{Lissajous Curve}:
    The Lissajous curve is a deterministic search pattern proposed by \cite{Steckenrider_Miller_Blankenship_Trujillo_Bluman_2024}, who found the method to be faster than traditional search patterns while still being comprehensive. In our implementation, we take linearly spaced intervals of the parameter $t$ to generate points at which the robot will make a scan. 
    
    \item \textit{Heuristic + Expanding Square}: This approach combines the proposed method with an expanding square pattern.
    
    \item \textit{Heuristic + Lissajous Curve}:  This approach combines the proposed method with a Lissajous pattern.

\end{itemize}

\subsection*{Search Performance Experiments}
\begin{figure}[h]
\centering
\includegraphics[width=\columnwidth]{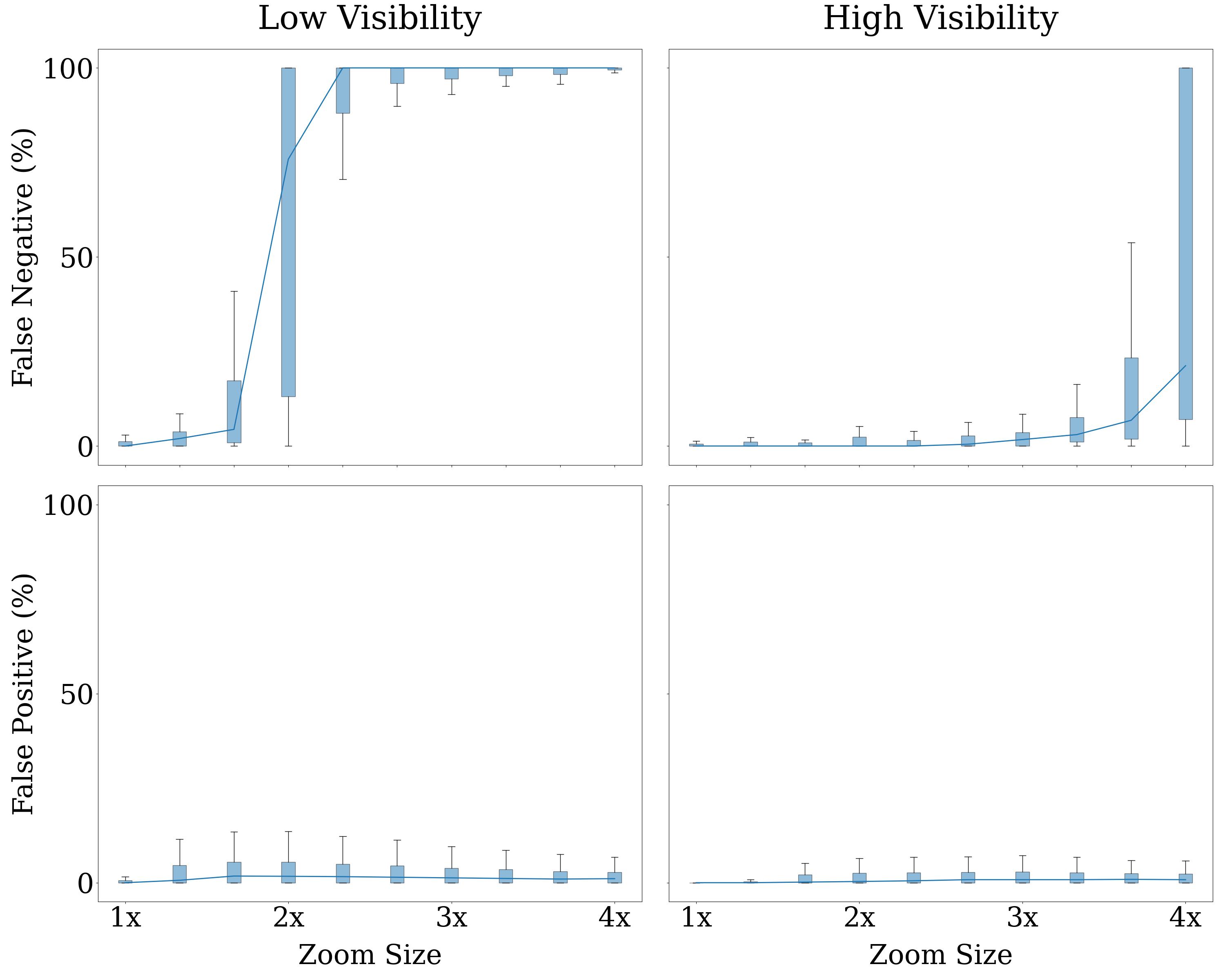}
\caption{The $y$ values correspond to the percentage of pixels that are misclassified. In a low visibility environment, nearly all of the soil samples would be missed at a zoom size of above 2x, corresponding to around 40\% between $h_{\max{}}$ and $h_{\min{}}$, and around 1-2\% of the remaining region would be detected as soil. On the other hand, in a high visibility environment, only a small portion of the soil samples will be missed in most cases, while the false positive rate remains at a similar level.}
\label{fig:visibility}
\end{figure}

We evaluated the performance of our algorithm under two scenarios: high environment visibility ($\theta = 0.7$), and low environmental visibility ($\theta = 0.2$). The detection accuracy for each test case is plotted in Figure \ref{fig:visibility}.

We run 5000 simulation trials for each test scenario and method, and we record the following metrics:
\begin{itemize}
    \item \textit{Number of Steps}: The number of steps in a trial before a robot finds a target. This roughly scales with the time taken, as each step comes with the time overhead of scanning, image processing, and decision making. 
    \item \textit{Distance Traveled}: The distance traveled by the agent in successful trials before finding a target sample. 
    \item \textit{Status}: Whether the robot finds a target. If it fails, the status includes the failure mode.
\end{itemize}

\section{RESULTS}
\label{sec:results}
From the experiments, our proposed method successfully demonstrated superior performance for all three metrics, and was robust to different visibility levels. 

\subsection*{High Visibility Experiment}
\begin{figure}[hbt!]
    \centering
    \includegraphics[width=0.99\columnwidth]{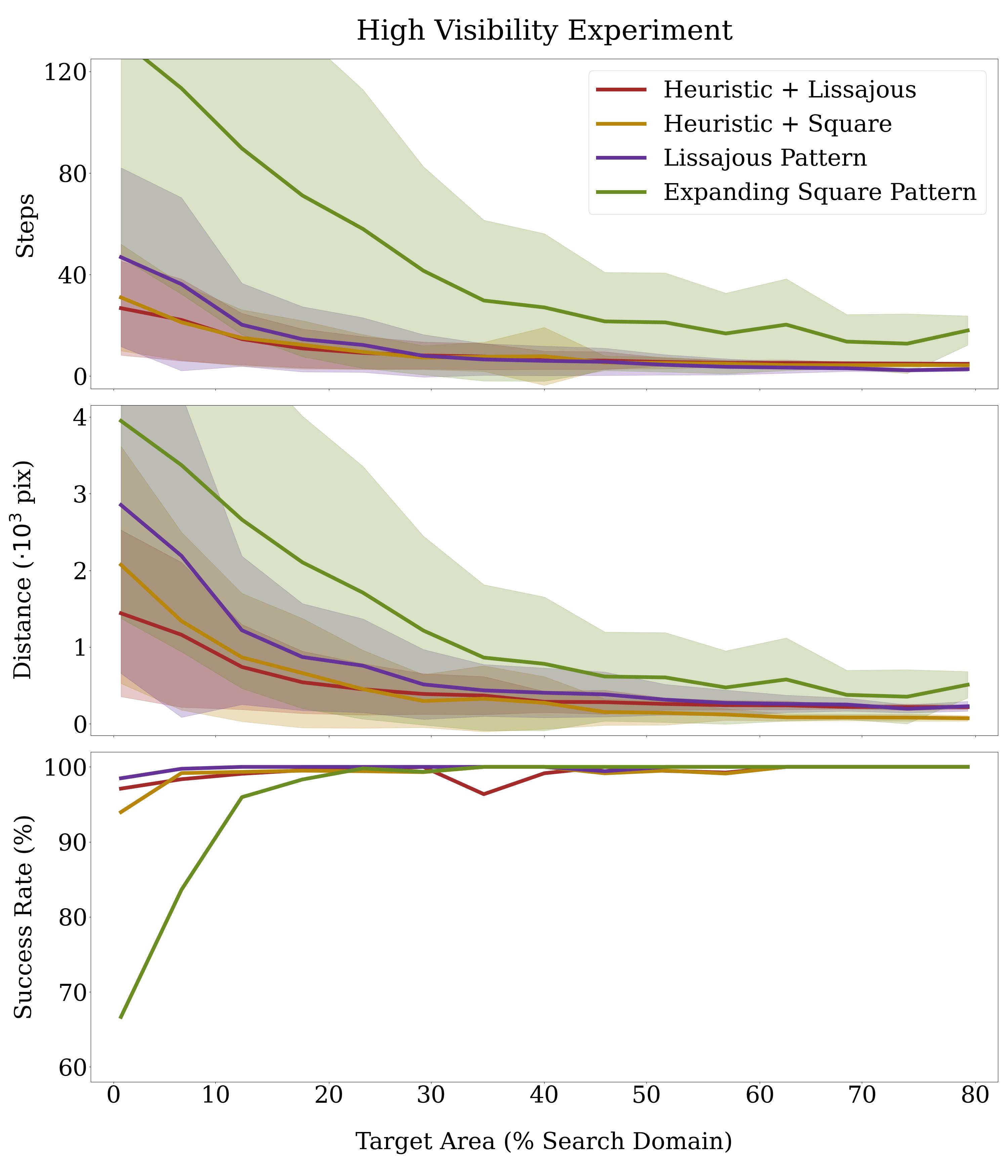}
    \caption{Among the four tested method, the Lissajous-based heuristic method outperforms all other methods in terms of step count and distance traveled while maintaining a near 100\% success rate across different soil abundance levels.}
    \vspace{-5pt}
    \label{fig:high_vis}
\end{figure}

\begin{figure}[hbt!]
    \centering
    \includegraphics[width=\columnwidth]{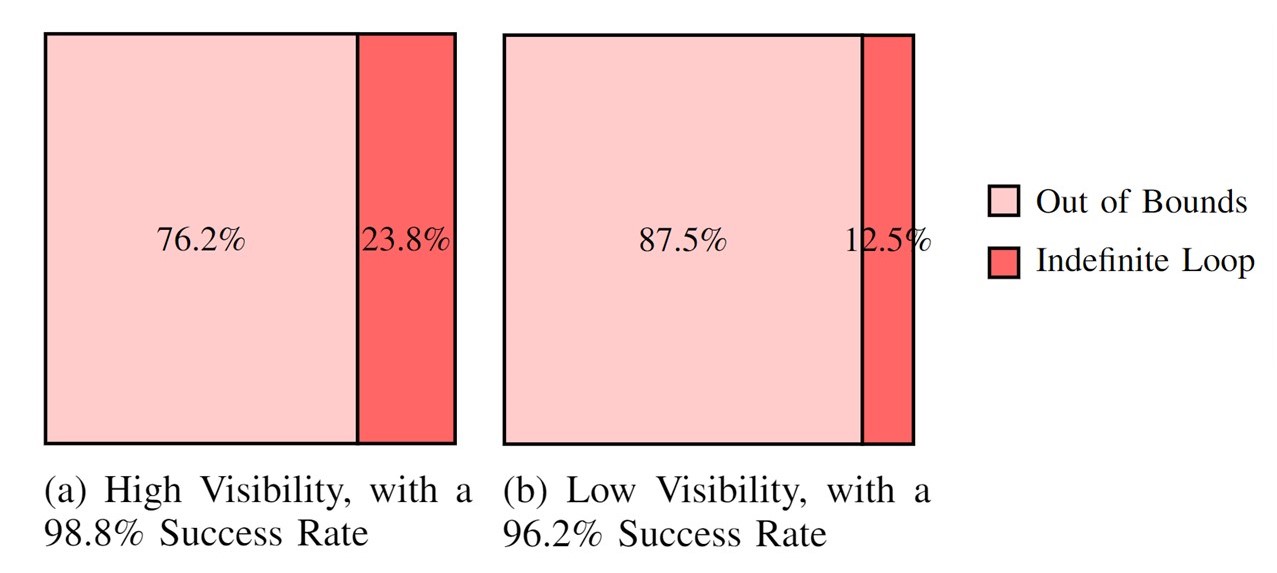}
    \caption{The proportions of each failure mode under different visibility conditions in environments with less than 15\% soil are shown. Out of bound failures are more common in low visibility conditions, which is caused by the higher probability of missed detection.}
    \label{fig:failure-results}
\end{figure}

In the high visibility case, we discovered that the heuristic guided search methods outperform the deterministic baseline pattern with respect to all three metrics, especially in environments with a sparse amount of soil, as shown in Figure \ref{fig:high_vis}. This outcome matched our expectations as the heuristic methods ideally would converge to the center of soil regions as soon as a small fraction of the soil is detected.
The heuristic methods with the two baselines performs similarly in terms of number of steps across the soil abundance spectrum, but the method with a Lissajous baseline took only around 80\% of traveled distance in sparse environments. In terms of success rate, the naive expanding square pattern experienced a significant number of failures in low soil abundance environments, which can be attributed to the large chosen step size that allows little to no overlap between two successive scans and causes smaller target areas that lie between to be omitted. On the other hand, the Lissajous-based method, both the naive and heuristic one, were able to achieve a success rate of above 98\% regardless of soil abundance. 
We identified two main failure modes of our proposed algorithm with a Lissajous baseline, indefinite looping and moving out of bounds:
\begin{itemize}
    \item In an indefinite looping scenario, the agent would spot a small soil region that is smaller than the minimum defined step size. As a result, the search algorithm will "overshoot" a step and the target would be in the direction which the agent came from, leading to an indefinite movement back and forth about the target region and a \textit{Truncated} final status.
    \item In an out of bounds case, the robot would fail to detect any soil regions along its search pattern, either due to false negative detection, or soil samples lying in between the gaps of two successive scans, leading to an \textit{Out of Bounds} final status.
\end{itemize}
Figure \ref{fig:failure-results} illustrates the proportion of each failure modes in a soil sparse environment.  
\begin{figure}[hbt!]
    \centering
    \includegraphics[width=0.99\columnwidth]{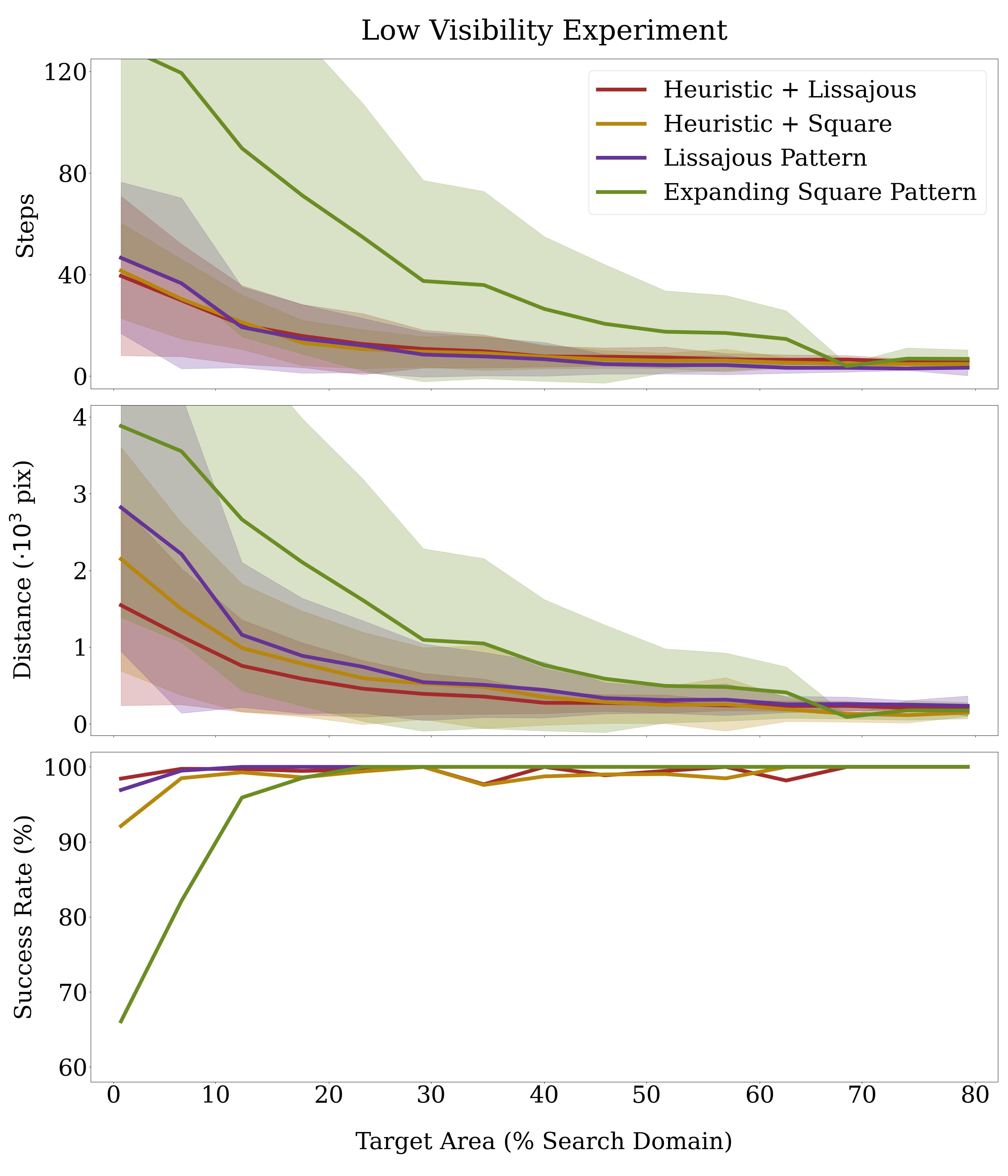}
    \caption{Similar to the high visibility case, the Lissajous-based heuristic method demonstrates superior performance compared to the other three methods.}
    \vspace{-7pt}
    \label{fig:low_vis}
\end{figure}
\subsection*{Low Visibility Experiment}
\begin{figure}[hbt!]
    \centering
    \includegraphics[width=\columnwidth]{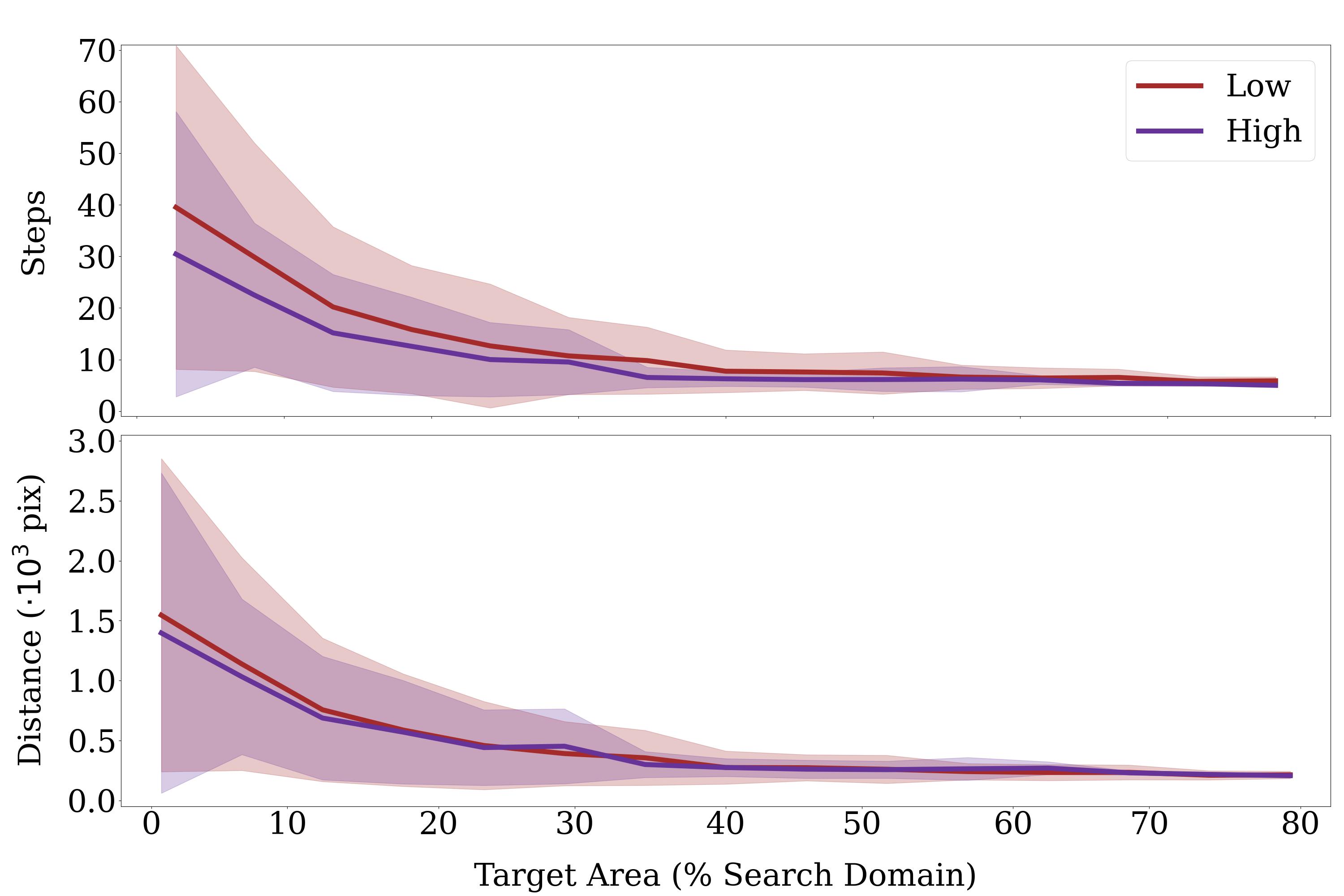}
    \caption{The performances of the Lissajous-based heuristic method under two different visibility levels. In environments with low soil level, the agent on average took only 10 more steps to locate targets with low visibility than with high visibility.}
    \vspace{-10pt}
    \label{fig:comp_vis}
\end{figure}

In the low visibility experiments, we discovered that the proposed heuristic still improved the performances of the baseline patterns despite the changed visibility, shown in Figure \ref{fig:low_vis}. The naive baselines performed similarly in this experiment as they did in the high visibility experiment. This matched our expectations as they were programmed to operate at the lowest height where noise has minimal effect. On the other hand, we found that visibility impairment only mildly degraded the performances of the heuristic methods. There was a slight success rate drop of 2.6\% for the Lissajous-based heuristic method, while the step number and total distance traveled only increased by a small margin, as shown in Figure \ref{fig:comp_vis}. This minor variation between the two experiments showcases the robustness of our proposed method against different visibility conditions with little hyper-parameter tuning required.

\section*{CONCLUSION}
We studied the problem of amorphous target search, which has application such as soil collection. We proposed a heuristic that drives the robot towards the center of a soil region using a camera with a limited field of view. Through simulation experiments, we verified that the proposed algorithm outperforms patterns used in search and rescue missions. The Lissajous-based heuristic search resulted in the least amount of steps and distance traveled across different soil abundance and visibility levels. 

This research can further be extended by incorporating frontier based exploration methods into our heuristic, and training the system with reinforcement learning to further improve resilience against visibility uncertainties. Lastly, an important future research direction is to further verify the effectiveness of this method through hardware experiments and field tests.

\bibliography{references} 
\bibliographystyle{ieeetr}

\end{document}